Daniel Hopp
Associate Statistician
Division on Globalisation and Development Strategies, UNCTAD
daniel.hopp@unctad.org

# Performance of LSTM Neural Networks in Nowcasting during the COVID-19 Crisis

## Abstract

The COVID-19 pandemic has demonstrated the increasing need of policymakers for timely estimates of macroeconomic variables. A prior UNCTAD research paper examined the suitability of long short-term memory artificial neural networks (LSTM) for performing economic nowcasting of this nature. Here, the LSTM's performance during the COVID-19 pandemic is compared and contrasted with that of the dynamic factor model (DFM), a commonly used methodology in the field. Three separate variables, global merchandise export values and volumes and global services exports, were nowcast with actual data vintages and performance evaluated for the second, third, and fourth quarters of 2020 and the first and second quarters of 2021. In terms of both mean absolute error and root mean square error, the LSTM obtained better performance in two-thirds of variable/quarter combinations, as well as displayed more gradual forecast evolutions with more consistent narratives and smaller revisions. Additionally, a methodology to introduce interpretability to LSTMs is introduced and made available in the accompanying *nowcast_lstm* Python library, which is now also available in R, MATLAB, and Julia.

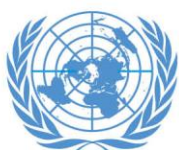

© 2021 United Nations

## Contents

### Acknowledgements

I would like to thank Anu Peltola her valuable comments and feedback.

# 1. Introduction

The COVID-19 pandemic wrought havoc on the global economy in 2020. In contrast with other economic crises, such as the 2008 financial crisis, there were not primarily macroeconomic factors at play, but rather epidemiological ones. As the threat of contagion forced innumerous business closures, especially in the service and tourism sector (UN, 2020), economic contraction followed. In order to combat these events, unprecedented in modern times, many governments implemented extensive stimulus measures to help people through the crisis. In the months following initial widespread global closures in March 2020, the importance of timely information on the state of national economies and the global economy became essential in quickly assessing both the impact of existing policy measures, as well in guiding future ones. The months long publication delays typical of many macroeconomic series, especially globally aggregated ones, such as GDP or international trade, were rendered even more of a barrier for guiding policy during such a quickly developing crisis (Gerhard et al., 2021).

In this scenario, nowcasting, the estimation of the current or near-current state of a target variable using information that is available more quickly, could be an essential tool in gaining insight to the COVID-19 pandemic's effect on the global economy. The COVID-19 pandemic proved a stress-test for existing nowcasting models, most having never before been confronted with such an extreme and dynamic crisis. These circumstances make 2020 a particularly interesting case in which to examine the performance of different nowcasting methodologies. This paper seeks to do just that, assessing two methodologies, the dynamic factor model (DFM), currently a popular choice in economic nowcasting, and the long short-term memory neural network (LSTM), explored in-depth in Hopp (2021).

Additionally, the dynamic economic situation naturally leads to much larger revisions in model predictions over time than would be expected in normal economic circumstances. This increases the value of causal inference into what is driving the change in a model's predictions. To that end, this paper also explores a methodology to introduce such causal inference to the outputs of the LSTM. This functionality has been added to the *nowcast_lstm* Python library, which is discussed in the relevant section 4.1. Finally, in order to further increase accessibility to the use of LSTMs in economic nowcasting, wrappers for R, MATLAB, and Julia for the *nowcast_lstm* library have been introduced, enabling the use of library from these languages without the need for Python knowledge. More information is available from the following locations:

- R: https://github.com/dhopp1/nowcastLSTM
- MATLAB: https://github.com/dhopp1/nowcast_lstm_matlab
- Julia: https://github.com/dhopp1/NowcastLSTM.jl

The remainder of this paper is structured as follows: the next section will provide more background information on nowcasting, including during the COVID-19 pandemic, and the LSTM methodology; section three will examine the relative performance of DFMs and LSTMs in nowcasting three series during the pandemic: global merchandise trade exports expressed in both values and volumes and global services exports; section four will introduce and examine a methodology for introducing causal inference to LSTM predictions, as well as introduce the wrappers for the *nowcast_lstm* library; section five will conclude and examine areas of further research.

____________________________________________________________________________________________

# 2. Background

## 2.1 Nowcasting in the context of the COVID-19 pandemic

Nowcasting is the forecasting of the current or near-current value of a variable, often using information that is published or made available more quickly than the variable of interest. Some commonly nowcasted series include GDP (Morgado et al., 2007; Giannone et al., 2009; Rossiter, 2010) and international trade (Cantú, 2018; Guichard and Rusticelli, 2011). These types of aggregated macroeconomic variables lend themselves well to the nowcasting paradigm, as they are often published later than some other economic indicators while still being of great interest to policymakers, investors, and firms. Some common methodologies to perform economic nowcasting include mixed data sampling (MIDAS) (Kuzin et al., 2009; Marcellino and Schumacher, 2010), dynamic factor models (DFM) (Guichard and Rusticelli, 2011; Corona et al., 2021), mixed-frequency vector autoregression (VAR) (Kuzin et al., 2009; Huber et al., 2020), and Bayesian vector autoregressions (Cimadomo et al., 2020). Hopp (2021) and Loermann and Maas (2019) examined neural networks' suitability to the application, more specifically long short-term memory (LSTM) networks in the case of the former. The LSTM methodology is explained further in section 2.2. For more information on nowcasting, including commentary on common data issues encountered in the field, see Hopp (2021) or Cimadomo et al. (2020).

Nowcasting became more relevant than ever in the wake of the economic fallout from the COVID-19 pandemic. Since March 2020, when many governments around the world began shutting down businesses and other forms of economic activity in response to the virus, transforming the crisis into a global one, the rate of change in the economy has been truly unprecedented (The World Bank, 2020). Furthermore, the epidemiological nature of the crisis and successive COVID-19 waves have meant that the economic recovery has not been one of monotonic recovery, as governments have often had to roll back and reinstate openings in response to the severity of local and national outbreaks. This has simultaneously increased the need for accurate, timely assessments of the economic situation to inform policy and mitigate economic impact on citizens, while making those assessments harder to acquire.

However, crisis often creates opportunity and breeds innovation, and the field of nowcasting has been no different. A wealth of papers relating to nowcasting during the COVID-19 pandemic have been published since March 2020. Many geographies are represented, including Canada (Chapman and Desai, 2021), Sub-Saharan Africa (Buell et al., 2021), the United States (Foroni et al., 2020), Mexico (Corona et al., 2021), and the Euro area (Huber et al., 2020), among others. Perhaps more interestingly, novel data sources have additionally been explored, for instance Google mobility data (Sampi and Jooste, 2020), retail payment system data (Chapman and Desai, 2021), Google search trends, and mobile payment data (Buell et al., 2021). Unfortunately, the longevity of the COVID-19 crisis to this point ensures that nowcasting its effects on the economy will remain fertile ground for new research in the coming months and years.

## 2.2 Long short-term memory neural networks

Having established the context in which the nowcasting exercise outlined in this paper takes place, this section will give a short background on the methodology employed. Artificial neural networks (ANNs) have risen in prominence in recent years due to their impressive performance in a variety of applications, including things like image classification and natural language processing. However, traditional feed-forward networks lack a temporal component, a frequent feature of many economic applications. The long short-term memory network architecture (LSTM) adds this component and renders them more suitable for application in the nowcasting context. For more information on how ANNs and LSTMs work, see: Hopp (2021), Singh and Prajneshu (2008), Sazli (2006), or Loermann and Maas (2019). For more detailed information on LSTMs' properties which make it suitable for nowcasting, see Hopp (2021), section 3.2.

# 3. Empirical analysis

## 3.1 Description of data and models

Hopp (2021) examined the LSTM's performance versus that of dynamic factor models (DFM) in nowcasting global merchandise and services trade. In that case, LSTMs were found to produce superior predictions. However, the test period was the fourth quarter of 2016 to the fourth quarter of 2019, a period when the target series' movements were much more muted than compared with 2020 and 2021. Furthermore, test performance was found using artificially simulated data vintages based on historical publication lags. The analysis performed in this paper seeks to build on those findings and further validate and stress test them with: A) a much more volatile and difficult to predict in context, and B) actual data vintages collected over the course of 2020 and 2021.

In this analysis, three target variables were again nowcast: global merchandise exports in both value (WTO, 2020) and volume (UNCTAD, 2021), and global services exports (UNCTAD, 2021). These are the same series examined in Hopp (2021). All target series were expressed in seasonally adjusted quarter over quarter growth rates. In total, 45 independent variables were used as inputs to estimate both a DFM and LSTM model for each target series: 17 for merchandise exports values, 17 for merchandise exports volumes, and 21 for services exports. Variables were sometimes used to estimate more than one target series. Input variables included things such as industrial production indices, manufacturing export order books, and retail trade indices, among others. See appendix 1 for a full list of input variables, including their geographies, frequencies, sources, and for which target series they were used. The same variables were used in estimating both the DFM and LSTM models to ensure maximum comparability. Input variables were a mix of monthly and quarterly frequencies expressed in period over period seasonally adjusted growth rates.

The DFM and LSTM models were trained on data dating from the second quarter of 2005 to the fourth quarter of 2019, representing the maximum extent of information a forecaster or policymaker would have had in the run up to the COVID-19 pandemic. Actual data vintages collected over the period from March 2020 to October 2021 were then used to assess model performance in nowcasting the target series from the second quarter of 2020 to the second quarter of 2021, an exceptionally volatile and difficult period to nowcast due to the unprecedented impacts on the global economy of the COVID-19 pandemic. Actual data vintages were collected on a monthly basis from March to July 2020, then on a weekly basis from August 2020 to October 2021.

The LSTM model used was the same examined in Hopp (2021), using the averaged output of 10 networks. For the logic of using the average of multiple networks' outputs, see Hopp (2021) sections 4.1 and 5, or Stock and Watson (2004). Hyperparameters were found by using the period from the second quarter of 2005 to the third quarter of 2016 as a training period, and the fourth quarter of 2016 to the fourth quarter of 2019 as a test period. Ragged edges were filled using the mean of each series, see Hopp (2021) section 3.2 for more information.

The DFM model used was that described in Cantú (2018), where a state-space representation is used to model the DFM under the assumption that the target and independent variables share a common underlying factor, as well as containing their own idiosyncratic component. Subsequently, the Kalman filter is applied and maximum likelihood estimation used to obtain parameter estimates. For more information on this specific DFM methodology, see Bańbura and Rünstler (2011) and Bok et al. (2018).

Once DFM and LSTM models were trained for each target series with data up until the fourth quarter of 2019, predictions could be obtained on actual monthly and weekly data vintages to see how the models' forecasts would have developed over time as the pandemic unfolded and its economic repercussions began to appear in the data. In this way, we can see what narratives and guidance the nowcasts would have provided to policy makers and analysts as well as assess their errors over time and final performance.

Predictions were made for each quarter on data vintages dating 100 days either forwards or backwards in time, to assess performance both early on, when little data for the period was available, and later on, when data on most independent series had been published.

## 3.2 Results

Figure 1 shows the development of the two models' predictions over time for the period from the second quarter of 2020 to the second quarter of 2021. The X axis shows the days difference from the target period. E.g., 0 days difference for 2020 Q2 refers to 1 June 2020, to 1 September for 2020 Q3, etc. The Y axis displays the quarter over quarter growth rate. The red line displays the actual observed growth rate, while the blue and green lines represent the predictions of the LSTM and DFM models, respectively. Each point making up the blue and green lines represents what the two models predicted the growth rate of the target series would be given the data available at that point in time. Generally, the predictions should move closer to the actuals line as time goes on and more data is released.

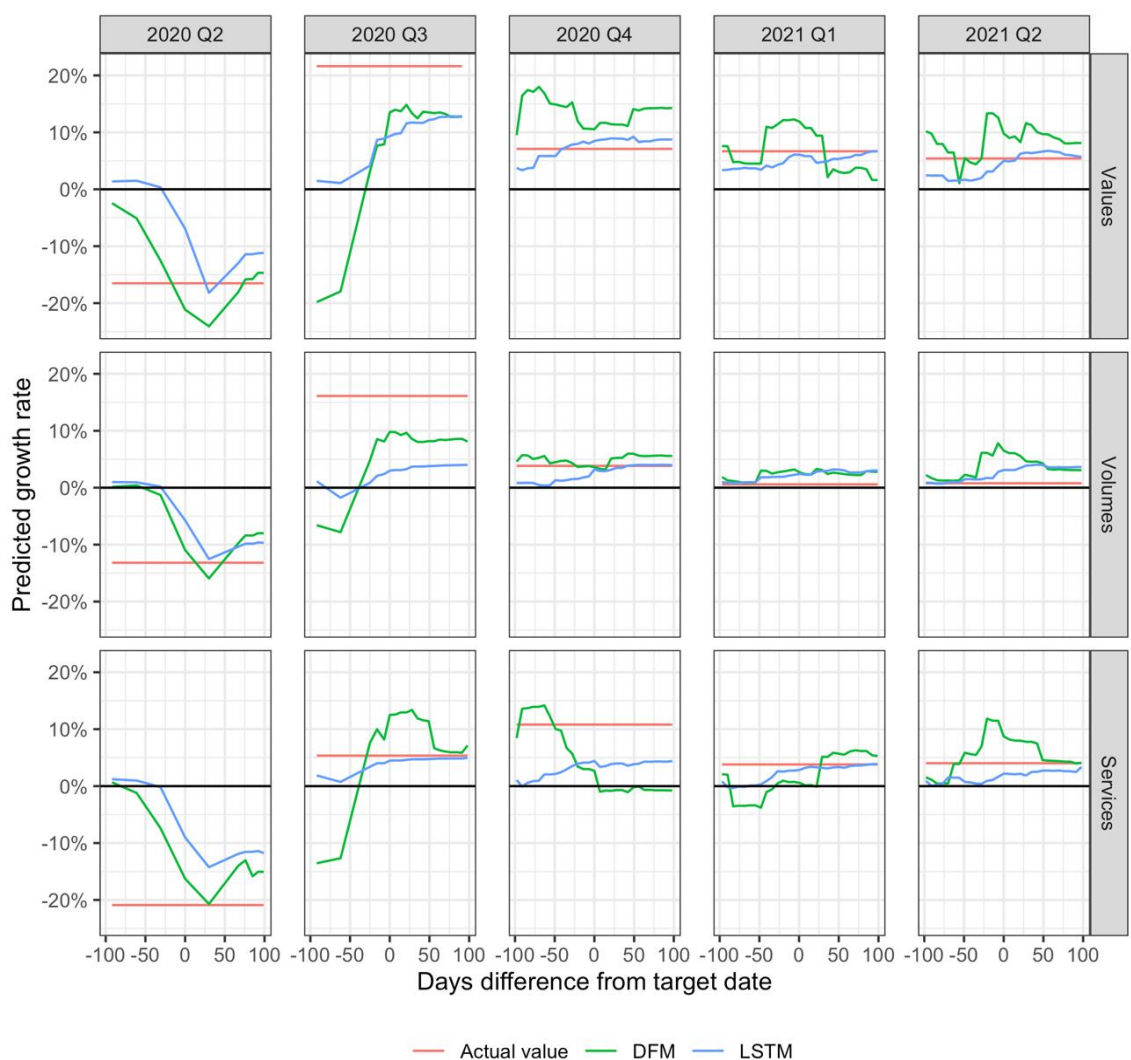

Figure 1. Nowcast evolution over time

*Note: For brevity, "Values" refers to global merchandise exports in values, "Volumes" refers to global merchandise exports in volumes, and "Services" refers to global services exports.*

**2020 Q2**
The first column of figure 1 details predictions for 2020 Q2. This was the first quarter where the full effects of the pandemic were reflected in economic data globally. While China was already experiencing lock downs in the first quarter of 2020, most other places did not until COVID-19 was declared a pandemic by the WHO on 11 March 2020 (WHO, 2020). The first quarter of 2020 was not assessed in this modelling exercise as UNCTAD did not begin the systematic gathering of actual data vintages until after this period had elapsed.

Global merchandise exports expressed in values dropped 16.5 per cent quarter over quarter in the second quarter of 2020. Between 2005 and 2021, this was the second largest decline recorded, second only to the fourth quarter of 2008, during the height of the financial crisis. While the DFM already began to pick up on contraction in March and April, it began severely revising its predictions downwards in May and June (day difference of 0 on the X axis), actually severely overshooting the eventual number by nearly 8 percentage points in July, as negative figures from April and May had been published, but not as many more positive ones from June had been. As data continued to be released through July, August, and September, it revised its predictions upwards before settling quite close to the actual figure in the beginning of September. The LSTM took longer to reflect the downturn, only heavily revising its predictions downwards in June. It displayed a similar shape to the DFM, with steep downward revision followed by upward correction. However, its post-July trough to peak delta was only 7 percentage points, compared with nearly 10 percentage points for the DFM.

Global merchandise exports expressed in volumes dropped by 13.2 per cent in the quarter, representing the largest decline recorded between 2005 and 2021, even greater than declines observed during the financial crisis. In this series, the DFM was slower to pick up on the decline compared with values, only revising predictions strongly downwards in June. Again, it overshot the mark and revised itself upwards after hitting its nadir in July. This time, however, it overshot the mark on the way up as well, and it finished predicting a decline that was only about 60 per cent as large as the actual observed decline. The LSTMs' predictions followed a similar pattern, declining sharply in June and July, then revising upwards afterwards. Its revisions, however, were significantly smaller than the DFMs', with a post-July trough to peak delta of only 3 percentage points, compared with the DFM's of 8 percentage points. Its final predictions also ended closer to the actual value.

Both models had a hard time picking up on the degree of decline for global services exports. Perhaps understandable, considering the series experienced its greatest decline in the second quarter of 2020 in the period from 2005 to 2021, almost doubling the next largest downturn experienced during the global financial crisis. Both models' predictions displayed similar shapes to the merchandise export series, with big revisions downwards followed by corrections. Again, the LSTM displayed smaller corrections, with a post-July trough to peak delta of 3 percentage points compared with the DFM's of 8 percentage points, though the DFM's final prediction was closer to the observed value.

**2020 Q3**
The third quarter of 2020, represented in the second column of figure 1, experienced strong recovery after astounding contractions in the second quarter. Though recovery had already begun in May and June of 2020, as summer in the northern hemisphere brought about partial economic reopening combined with adaptation to the circumstances of the pandemic, it was visible in earnest in the third quarter.

Global merchandise exports expressed in values ended up growing an impressive 21.6 per cent quarter over quarter in the third quarter of 2020, albeit from the low base of the second quarter. In June and July of 2020, the DFM was still forecasting very negative growth, as there was little indication in the data that robust recovery was on the horizon. By August 2020, the DFM finally began revising its forecasts upwards, reaching a high of 14.8 per cent before more or less stagnating and finishing at 12.8 per cent. The LSTM followed a more gradual path to the same conclusion as the DFM, starting out forecasting 1.5 per cent growth in June 2020, gradually building towards a final forecast of 12.8 per cent, quite similar to the DFM. Like the DFM, the LSTM experienced its biggest upwards revision in August, as it became clearer in the data that the poor economic conditions of the second quarter would not continue into the third.

Global merchandise exports expressed in volumes grew 16.1 per cent quarter over quarter in the third quarter of 2020. Both the DFM and the LSTM significantly underestimated this growth, especially the LSTM. The DFM displayed a similar pattern to that observed in merchandise values, starting off quite negative in June and July, before revising upwards in August and remaining more or less stagnant subsequently. The LSTM followed a similar pattern but ended up forecasting about half the growth of the DFM, ending at 4 per cent and 8.1 per cent, respectively.

Global services exports experienced a much more muted recovery in the third quarter, growing only 5.3 per cent quarter over quarter. This is most likely due to the nature of the COVID-19 pandemic and its outsized impact on services-oriented activities, such as events and dining. The DFM again started out forecasting substantial contraction in June and July before revising itself upwards starting in August. This time, however, it substantially overshot the mark, maxing out with a prediction of 13.4 per cent quarter over quarter growth at the end of September before revising itself significantly downwards at the end of October, ending with a forecast of 7.1 per cent. The LSTM again displayed a more gradual forecast development, starting out predicting minimal growth in June and July before beginning to gradually revise itself upwards in August. By September, the forecast had reached 4.7 per cent, where it would more or less remain until the end of the prediction period, finishing with a forecast of 5.0 per cent, remarkably close to the actual final observed value.

**2020 Q4**
The fourth quarter of 2020, the third column in figure 1, saw the recovery continue, but not without its complications. Second and third COVID-19 waves, for instance in Europe and the United States in October and November 2020, saw the reintroduction of lockdown measures. The result was significantly slower global merchandise exports quarter on quarter growth. Global services exports actually grew at a faster clip in the fourth quarter than in the third quarter, partially due to the fact that the third quarter recovery was so anemic in comparison to the recovery in merchandise exports, coupled with stronger declines in both the first and second quarters compared with merchandise trade.

Global merchandise exports expressed in values ended up growing 7.1 per cent quarter over quarter in the fourth quarter of 2020, constrained partially by subsequent COVID-19 waves in the northern hemisphere's autumn and winter, and partially due to the strong growth already accrued in the third quarter. The DFM consistently overestimated the rate of growth over the course of the prediction period, maxing out with a forecast of 18 per cent growth at the end of September and bottoming out with a forecast of 10.6 per cent growth in the beginning of December, before creeping up again to 14.3 per cent by March 2021. The LSTM similarly overestimated growth in the fourth quarter, though not by nearly as much. The forecast started below the observed actual, forecasting 4 per cent growth until the end of September, after which it slowly revised itself upwards before settling around 8 per cent the beginning of November. The forecast stayed near that level until the end of the forecasting period, finishing at 8.7 per cent in March 2021.

Global merchandise exports expressed in volumes grew just 3.8 per cent quarter over in the fourth quarter of 2020. Compared with other periods, the DFM was relatively consistent in its predictions over the course of the prediction period, hovering around 5 per cent the majority of the time. The predictions did dip to around 3 per cent in mid-December 2020 but rose again to finish at 5.6 per cent. The LSTM again displayed its typical pattern of starting with low or conservative estimates, and gradually building towards a final value, in this case 4.0 per cent, very close to the actual.

Global services exports experienced comparatively robust quarter over quarter growth of 10.8 per cent in the fourth quarter of 2020. However, once the relatively low growth rate of the third quarter is taken into account, the feat appears less impressive. The DFM did a poor job of picking up this continued fourth quarter growth, beginning with optimistic forecasts in September before continually revising them downwards until December, settling at around -0.1 per cent growth at that time and remaining there until the end of the prediction period. The LSTM, on the other hand, once again displayed the common pattern visible in all post-Q2 2020 quarters, that of beginning conservatively and gradually building towards a final prediction. The LSTM's predictions started at around 1-2 per cent, before reaching 3 per cent by the end of October. Predictions would stay around that range, finishing the prediction period at 4.4 per cent, markedly closer to the observed actual than the DFM, which actually predicted contraction by the end of the prediction period, though still significantly underestimating the eventual observed value.

**2021 Q1**
The first quarter of 2021, the fourth column in figure 1, saw the beginning of large-scale vaccination campaigns in several large, developed regions such as the United States of America and Europe (Our World in Data, 2021). Additionally, as caseloads had achieved a local peak in many regions either in the fourth quarter or the very beginning of the first quarter, global cases actually declined for the first two months of the quarter, before creeping up again in March (WHO, 2021). Even still, after two months of strong growth, the first quarter of 2021 saw growth rates decline from the fourth quarter for all three target series.

Global merchandise exports expressed in values grew 6.7 per cent in the first quarter of 2021, lower than its 7.1 per cent pace in the fourth quarter. The DFM began the prediction period slightly underestimating growth, forecasting around 4.5 per cent until mid-January 2021, at which point it significantly revised itself upwards to 11 per cent. It would remain at this higher point until April, when worse-than-expected Q4 actuals revised the forecast strongly downwards. From that point onwards it would underestimate growth, finishing the prediction period with a forecast of 1.6 per cent. The LSTM began the prediction period at 3.4 per cent, slowly revising itself upwards over the course of the prediction period to finish at 6.7 per cent, remarkably, mirroring the actual observed value up to three decimal places. The LSTM's forecast range over the entirety of the prediction period was 3.3 percentage points, compared with 10.6 percentage points for the DFM.

Global merchandise exports expressed in volumes grew 0.6 per cent in the first quarter of 2021, compared with 3.8 per cent in the fourth quarter. In contrast to the DFM's first quarter predictions for values and services, its predictions for volumes remained rather consistent throughout the prediction period. It began the period with predictions of 1.9 per cent growth and ended predicting 2.8 per cent growth, with some up and down variation in between. The LSTM told largely the same story, beginning the period predicting 1 per cent growth and ending with a prediction of 3 per cent growth, corresponding closely to the DFM's predictions throughout the prediction period.

After robust growth in the fourth quarter, global services exports grew a more measured 3.8 per cent in the first quarter of 2021. The DFM began the quarter predicting contraction of around 3.5 per cent until mid-January 2021, at which time it significantly revised itself upwards to growth of around 0 per cent. Predictions would stay at that level until April, at which time they were revised upwards again to between 5 and 6 per cent, finishing with a prediction of 5.3 per cent. The LSTM also began the period predicting contraction, albeit of less than 1 per cent. By February, the forecast had reached 2.6 per cent and would end the prediction period with a prediction of 3.9 per cent, quite close to the actual value, with minimal variation in between.

**2021 Q2**
During the second quarter of 2021, the final column in figure 1, vaccination campaigns continued and picked up pace, with issues surrounding uptake mostly limited to supply. Despite the increasing availability of vaccines and warmer weather in the northern hemisphere, all three series experienced very similar levels of growth to the first quarter.

Global merchandise exports expressed in values grew 5.4 per cent in the first quarter of 2021, a bit slower than the 6.7 per cent managed in the first quarter. The DFM struggled to maintain a consistent message during the prediction period, beginning with a prediction of more than 10 per cent growth, revising quickly down to just 1 per cent by April, then moving quickly up again, maxing out at 13 per cent, before moving down to 8 per cent, then up again to 11 per cent, before finally settling around 8 per cent. The LSTM began the period with predictions hovering between 1 and 1.5 per cent, before rising to 5 per cent over the course of several weeks in May. It would then hover between 5 and 6 per cent for the rest of the prediction period, settling at 5.8 per cent.

Global merchandise exports expressed in volumes grew 0.8 per cent in the first quarter of 2021, compared with 0.6 per cent in the fourth quarter. Both models started off with forecasts near to the eventual actual figure before revising themselves upwards to above 3 per cent, thus overshooting. The DFM went as high as 8 per cent at the end of May before falling back. At the same period of May, the LSTM was still predicting growth of under 2 per cent. From May onwards, its predictions would remain relatively constant between 3 and 4 per cent.

Global services exports grew 4.0 per cent in the second quarter of 2021, very similar to the 3.8 they grew in the first quarter. The DFM began the prediction period with forecasts of between 0 and 1 per cent growth, before rising as high as 11.5 per cent by the end of May. It subsequently fell quickly to 8 per cent and remained there until mid-July, before falling quickly again to around 4 per cent, where it would remain until the end of the prediction period. The LSTM's predictions hovered between 0 and 1.5 per cent growth until the end of May, at which time they grew to hover between 2 and 3 per cent, ending at 3.4 per cent.

Tables 1, 2, and 3 display the mean absolute error (MAE) and root mean square error (RMSE) of the two models over the entire prediction period spanning from 100 days before the target period to 100 days after the target period. The values can be interpretated as the average deviation of the blue and green prediction lines from figure 1 from the red actuals line, expressed in either absolute or squared deviation. A one-tailed t-test was performed on all DFM and LSTM errors over the prediction period for each quarter with the alternative hypothesis that the LSTM errors were lower. Results are displayed in the LSTM columns.

Table 1. Average performance metrics, global merchandise trade exports, values

| Period | DFM MAE | LSTM MAE | DFM RMSE | LSTM RMSE |
|---|---|---|---|---|
| Q2 2020 | 0.0483 | 0.0884 | 0.0658 | 0.107 |
| Q3 2020 | 0.1277 | 0.1171 | 0.1616 | 0.1226 |
| Q4 2020 | 0.0654 | 0.0163*** | 0.0693 | 0.0184*** |
| Q1 2021 | 0.0347 | 0.0176*** | 0.0372 | 0.0207*** |
| Q2 2021 | 0.0356 | 0.0194*** | 0.0413 | 0.0234*** |

Note: *p<0.05 **p<0.01 ***p<0.001

Table 2. Average performance metrics, global merchandise trade exports, volumes

| Period | DFM MAE | LSTM MAE | DFM RMSE | LSTM RMSE |
|---|---|---|---|---|
| Q2 2020 | 0.0671 | 0.0663 | 0.0792 | 0.083 |
| Q3 2020 | 0.0927 | 0.1314 | 0.1043 | 0.1321 |
| Q4 2020 | 0.0119 | 0.015 | 0.0136 | 0.0196 |
| Q1 2021 | 0.0174 | 0.0156 | 0.0189 | 0.0176 |
| Q2 2021 | 0.029 | 0.0172** | 0.0346 | 0.021** |

Note: *p<0.05 **p<0.01 ***p<0.001

Table 3. Average performance metrics, global services exports

| Period | DFM MAE | LSTM MAE | DFM RMSE | LSTM RMSE |
|---|---|---|---|---|
| Q2 2020 | 0.091 | 0.1295 | 0.1122 | 0.1418 |
| Q3 2020 | 0.0547 | 0.011*** | 0.075 | 0.0153* |
| Q4 2020 | 0.0766 | 0.0767 | 0.087 | 0.0778 |
| Q1 2021 | 0.036 | 0.0158*** | 0.0416 | 0.0216** |
| Q2 2021 | 0.0266 | 0.0233 | 0.0349 | 0.025* |

Note: *p<0.05 **p<0.01 ***p<0.001

In terms of both MAE and RMSE, over the development of the whole prediction period, the LSTM provided better estimates on average in 10 of the 15 period-target series combinations, with statistical significance in 6 of those cases for MAE and 7 for RMSE.

## 3.3 Comparison with DFM

Before further comment on the performance and prediction qualities of the LSTM versus the DFM is made, it should be reiterated that the series mean was used to fill in missing values in ragged edges for the results obtained in figure 1 and tables 1, 2, and 3. Using the mean as opposed to ARMA estimations for filling ragged edges, the other option available in the *nowcast_lstm* library, naturally has some impact on the development of the LSTM's predictions, especially early in the prediction period. For insight on how using ARMA would have impacted the LSTM's predictions, refer to appendix 2, which shows the same information as figure 1 with the addition of LSTM ARMA ragged edge filling predictions. Using ARMA increases the reactiveness of the LSTM forecast, most evident early in the 2020 Q3 prediction period, but the findings and conclusions drawn from the previous section and below largely hold true for both the mean and ARMA ragged edge filling approaches.

Observation of the development of the two methodologies' predictions during the COVID-19 pandemic leads to two broad conclusions of these methodologies on this dataset. The first, that the DFM is generally more reactive, and second, that the DFM is much more influenced by previous values of the target variable, especially early on in the prediction period. It could be argued that both of these observations make the LSTM's predictions more suitable for nowcasting during economic disruptions, like those caused by the COVID-19 pandemic, or any other exceptionally volatile context.

The first observation, regarding the DFM's increased responsiveness, means that the DFM will respond more quickly and to a greater degree to changes in the data. At first glance, this may seem to be an advantage in the COVID-19 context. Indeed, considering only the second quarter of 2020, the DFM did generally perform better over the whole prediction period. It responded to the negative indications in the data significantly more quickly than the LSTM. However, this behavior is more beneficial when strong signals in the data remain consistent throughout the period. Even within the second quarter, extremely negative signals in the data were not distributed evenly, and this is visible in the development of the two models' predictions over the period. Extreme contractions recorded in April and the first half of May 2020 were followed by strong growth in June, as restrictions were lifted. As a result, both models had to revise their forecasts upwards in the latter half of the prediction period. In all three cases, the DFM, as the more reactive model, had to revise its predictions more heavily than the LSTM. For instance, in the case of merchandise export values, by July 2020 the DFM had significantly overshot the degree of contraction, predicting growth 7.6 percentage points lower than the actual. At the same time period, the LSTM was predicting growth just 1.7 percentage points lower than the actual. The LSTM, not having reacted as strongly or quickly to April and May signals, had more slack than the DFM to correct course in later weeks without the need for strong revisions.

This behavior is apparent in subsequent quarters as well, where the DFM reacts to the same data releases with much stronger revisions. Figure 2 quantifies this observation, displaying the share of weeks or data releases in each quarter where either the DFM or LSTM had a bigger revision to its forecast.

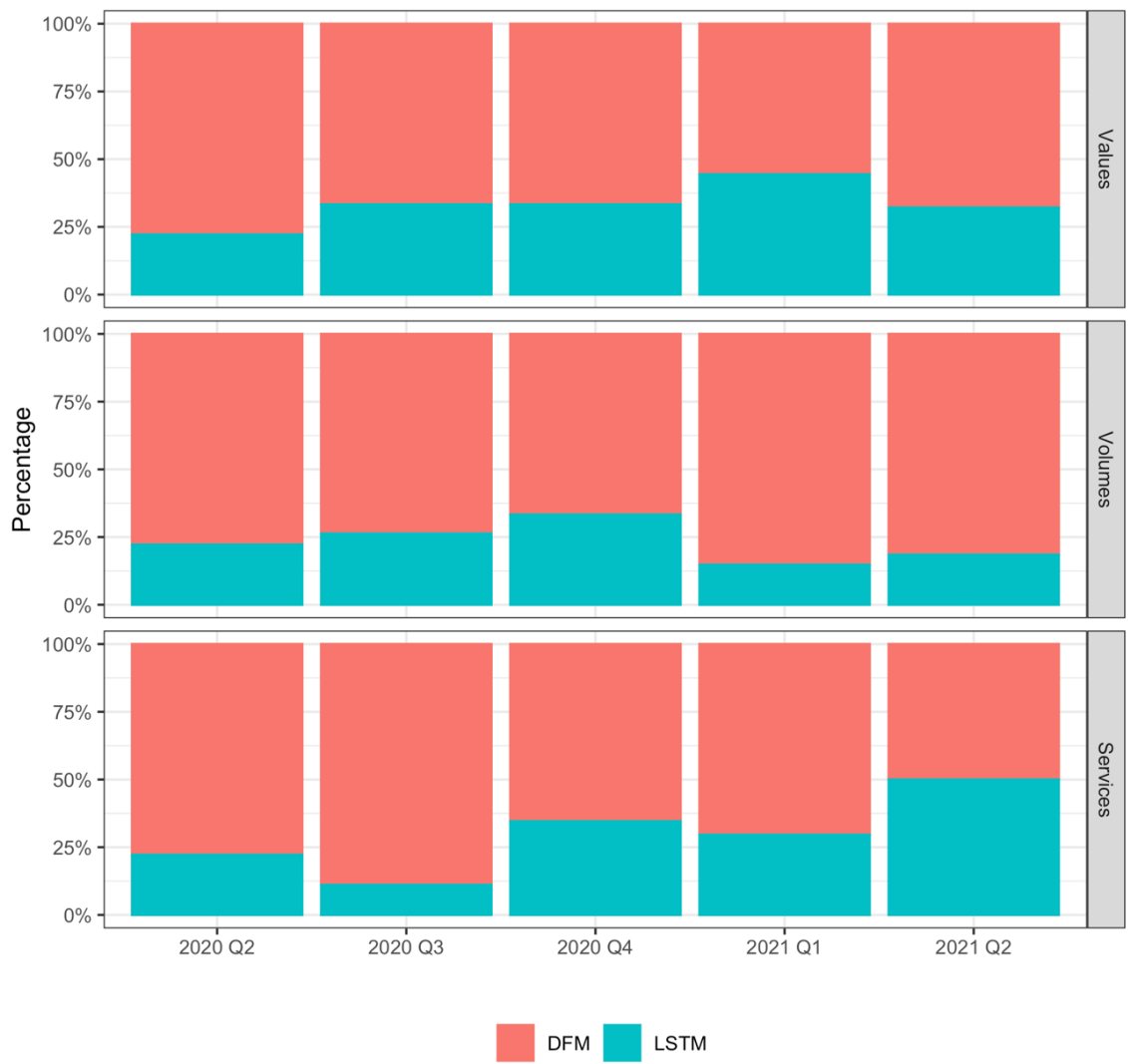

Figure 2. Share of weeks with a bigger revision

In every quarter-target series combination but one, the DFM more often had the bigger week to week revision than the LSTM, sometimes drastically so, as in services in the third quarter of 2020, where the DFM had a bigger week to week revision during 89 per cent of the prediction period. The two models only achieved parity in 2021 Q2 for services. Figure 3 further reinforces the observation, displaying the average weekly revision's absolute value for the two methodologies by target variable and quarter. The LSTM had smaller revisions on average for every target-quarter combination, often drastically so. By this metric, even target-periods which look favorable to the DFM in figure 2, such as 2021 Q1 values or 2021 Q2 services, do not compare well. The average weekly revision for the LSTM in these two periods is just one fifth and one third that of the DFM, respectively. This suggests that in weeks where the DFM has a smaller revision, it is smaller than an also small LSTM revision, whereas the DFM often has very large revisions which are not matched by the LSTM. In short, in this analysis, with a given data release, the DFM is more likely to revise its predictions much more heavily than the LSTM.

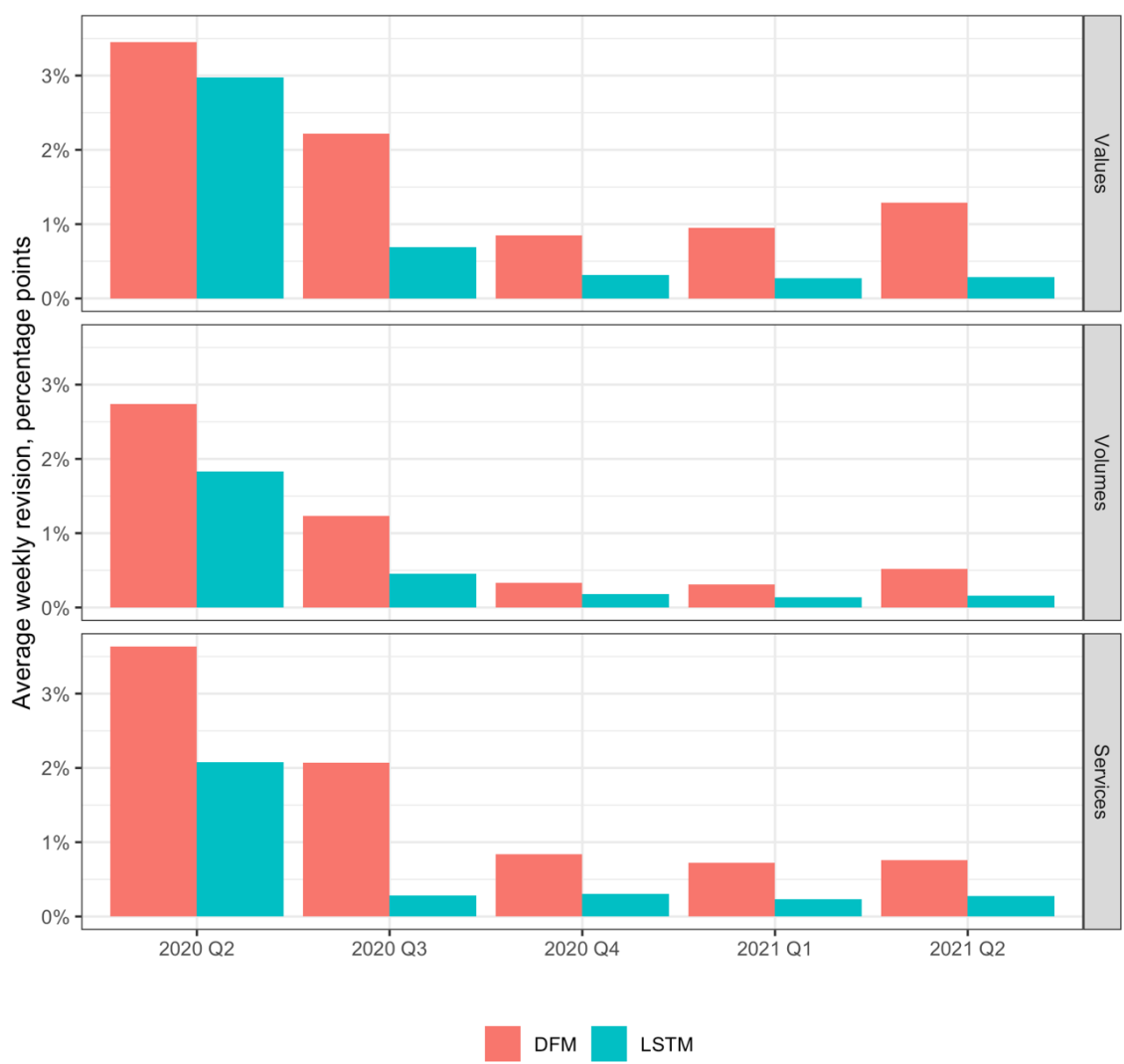

Figure 3. Average weekly revision

If a methodology produces predictions that are liable to be drastically altered not only over the course of the prediction period, but even from week to week, that limits its usefulness and the degree to which decisions can be made from its outputs. Services in the fourth quarter of 2020 are a good illustration of the contrasting nature of the two methods' prediction development. In the first week of the prediction period, the DFM started out forecasting very strong growth of 8.4 per cent. The next week, it revised its predictions upwards to 13.6 per cent. It continued to revise itself even higher the next four weeks, before reversing trend and continually revising downwards before plateauing at around 3 per cent. Then, one week in December, it revised itself down by 3 percentage points, beginning to forecast a contraction of about 1 per cent until the end of the prediction period. Over the same period, the LSTM began by forecasting modest growth of 1.1 per cent. It slowly revised itself upwards for the next two months, hitting 4 per cent by the beginning of November, before remaining at that level for the rest of the prediction period. This relationship and pattern are broadly generalizable to all of the quarter-target series combinations examined in the analysis.

This leads to the second main observation, that the DFM is more heavily influenced by previous values of the target variable. The LSTM does not take into account previous values of the target variable when making its predictions. The DFM does, however, due to its architecture, where the target variable is also used in estimating the latent underlying factor(s) (Bok et al., 2018). Especially early in the prediction period, we can see how the DFM's predictions are heavily influenced by either the previous value or the previous prediction of the target variable, observable in figure 1 as its predictions from one quarter appear to "flow into" those for the next. The LSTM does not display this behavior. Under normal circumstances, this characteristic is not particularly detrimental, as the difference between growth rates quarter to quarter does not widely differ historically, and early in a prediction period the prior observation would have been a sensible place for a forecast to start. For instance, between 2011 and 2019, the median absolute value of the difference between a quarter's growth rate and the previous quarter's was just 1.6, 0.8, and 1.5 percentage points for values, volumes, and services, respectively. Under the extraordinarily volatile conditions during the COVID-19 pandemic, however, where these quarter-to-quarter differences are more than an order of magnitude higher, this severely hampers the usefulness and accuracy of the DFM's early predictions. Because there was a sharp contraction in the second quarter of 2020, the DFM predicted similar results early in the prediction period for the third quarter. It was only able to generate more accurate predictions once more data were released signaling a strong reversal in the trend. Similarly, early in the fourth quarter prediction period for merchandise values, it again predicted extremely strong growth, as that was observed in the third. However, it was highly unlikely that such an exceptional growth rate would be maintained in the fourth quarter, something it could only reflect much later in the prediction period. The LSTM in all cases started with much more conservative estimates, which then gradually built towards a final prediction as more data were released. This behavior could be desirable in a nowcast during normal periods, but perhaps especially so in a volatile context such as the COVID-19 pandemic, where early estimates are ideally conservative to leave room for uncertainty, slowly building towards a final prediction as more data are released and confidence grows. This initial conservatism is enhanced in the LSTM by using mean-filling for ragged edges as opposed to ARMA, which displays prediction evolution slightly closer to the DFM's (see appendix 2). Even in the ARMA case, however, initial predictions remain much more conservative than the DFM's and later revisions resemble the mean-filling approach's very closely.

# 4. Library extensions

## 4.1 Model interpretability and news

One of the primary limitations of the LSTM approach identified in Hopp (2021) was the inability to gain insight on the impact of new data releases on the model's predictions. Discussion on how to obtain this information with the DFM is available in Bok et al. (2018), and a live application is available on the New York Fed's website for nowcasting US GDP growth (Federal Reserve Bank of New York, 2021). In release v0.1.5 of the *nowcast_lstm* library (Hopp, 2020), functionality was added that enables the generation of data releases' contributions to the LSTM's predictions. The methodology employed is similar to a simplified implementation of calculating Shapley values, whose use in adding interpretability to machine learning methodologies is explained in further detail in section 5.9 of Molnar (2019).

Given two data releases, week 1 and week 2, the contribution of new data releases and data revisions to the revised prediction between the two weeks is calculated as follows: for each variable with new data available in week 2, that new data is withheld and replaced with week 1's value. The change in the model's prediction is recorded as that
___

variable's contribution to the change in the prediction. Once the process is repeated for all variables with new data, predictions are obtained using data from week 2, but with the same missing values or ragged edges as were present in week 1. In this way, the contribution of data revisions, and not just new data releases can be ascertained. Once contributions for all new data releases and revisions are obtained, the values are rescaled to equal the delta between week 1 and week 2's predictions. In empirical testing, this rescaling factor is almost always close to 1.

This method should be considered experimental, and further research is necessary to examine additional and more robust ways of implementing interpretability within the LSTM framework. Further information on usage and methodology is available on the repository's page: https://github.com/dhopp1/nowcast_lstm.

## 4.2 R, MATLAB, and Julia wrappers

To further increase accessibility to the LSTM methodology examined in this paper, R, MATLAB, and Julia wrappers have been developed for the Python library. Python does need to be installed on the user's system; however, no Python knowledge is required to obtain full functionality from the library, all commands can be run directly from R, MATLAB, or Julia. Additional information as well as example files are available on the repositories' pages.

- R: https://github.com/dhopp1/nowcastLSTM
- MATLAB: https://github.com/dhopp1/nowcast_lstm_matlab
- Julia: https://github.com/dhopp1/NowcastLSTM.jl

# 5. Conclusion

The COVID-19 pandemic has made clearer than ever the need for timely statistics and estimates of the state of the economy. Engendering near unprecedented developments and changes in the global economy, both in terms of pace of change and degree of change, the pandemic has been a significant stress test for all types of economic accounting, forecasting, and modeling (European Commission, 2020; Gerhard et al., 2021; Pohlman and Reynolds, 2020). Nowcasting, which leverages the information contained in timelier variables to produce real-time estimates of variables published with long delays, is well-placed to address this need.

In this context, and with the benefit of five quarters of actual data vintages gathered during the crisis, this paper has sought to assess the performance of the LSTM neural network architecture versus an implementation of the widely adopted DFM methodology in nowcasting global merchandise and service exports. Further validation of the performance of a novel methodology in the LSTM could help enrich policy-makers' toolboxes and better equip them in dealing with and quantifying the next economic crisis.

Findings from the empirical analysis were encouraging for the adoption of the LSTM architecture. Five quarters, dating from the second quarter of 2020 to the second quarter of 2021, were nowcast for three target variables, seasonally adjusted quarter over quarter growth rates of global merchandise export values and volumes and global services exports, over a period of 100 days preceding and succeeding the end of each quarter. Of the 15 quarter-series combinations, the LSTM's predictions had a lower MAE and RMSE in 10 of them. This is evidence for, at the very least, competitive performance of the LSTM in comparison with the DFM, if not superior performance, depending on application requirements.

Beyond average performance over the prediction period, the LSTM's predictions were found to be more stable and less volatile, with the DFM registering larger week to week revisions in response to new data releases than the LSTM 70 per cent of the time across all quarter-series combinations. The LSTM's revisions were in turn on average only one third as big as the DFM's across all quarter-series combinations. The LSTM's predictions were found to most often follow the pattern of beginning with conservative estimates that slowly built towards a final prediction, with few radical course corrections or revisions, which were often observed in the DFM's predictions. This resulted in a less reactive model than the DFM, with the LSTM for instance slower to pick up on the large declines observed in the second quarter of 2020. However, this characteristic can also be viewed as a desirable feature in a forecast during volatile times, when extreme values are not guaranteed to continue in either direction or magnitude and there is a high degree of fluctuation and volatility.

Despite encouraging results from this analysis and that in Hopp (2021), research should continue into the use of the LSTM methodology for economic nowcasting, for instance with different target series, over different time periods and frequencies, and with different and higher-frequency explanatory variables. To further facilitate adoption and research, R, MATLAB, and Julia wrappers have been developed for the *nowcast_lstm* Python library (Hopp, 2021b). Additionally, functionality has been added to the library to enable a degree of model interpretability using simplified versions of Shapley values, with more information available in Hopp (2020). Hopefully, these resources and findings encourage continued work on LSTMs in the nowcasting context.

___________________________________________________________________________

# Appendix

## Appendix 1. Variables used in model estimation

*Note: For brevity, "values" refers to global merchandise exports in values, "volumes" refers to global merchandise exports in volumes, and "services" refers to global services exports.*

| Variable | Geography | Frequency | Source | Used to predict |
|---|---|---|---|---|
| Business confidence index | Netherlands | monthly | OECD | values, volumes, services |
| Business confidence index | Japan | monthly | OECD | values |
| Construction index | Canada | monthly | OECD | values |
| Consumer confidence index | Brazil | monthly | OECD | services |
| Container throughput index | global | monthly | RWI/ISL | volumes |
| Export prices of manufactures | global | monthly | WTO | services |
| Export volume of goods and services | Germany | quarterly | OECD | services |
| Export volume of goods and services | United States | quarterly | OECD | values, services |
| Export volumes | Eastern Europe and CIS | monthly | CPB | values |
| Export volumes | emerging Asia | monthly | CPB | volumes |
| Export volumes | Euro Area | monthly | CPB | values, services |
| Export volumes | global | quarterly | UNCTAD | volumes |
| Export volumes | Japan | monthly | CPB | volumes |
| Exports of services | EZ19 | monthly | ECB | values |
| Exports of services | global | quarterly | UNCTAD | services |
| Exports of services | Japan | monthly | BOJ | volumes |
| Exports of services | Singapore | quarterly | IMF | services |
| Exports of services | United States | monthly | FRED | services |
| GDP volume | United States | quarterly | OECD | services |
| Industrial production index | EU27 | monthly | Eurostat | services |
| Industrial production index | Germany | monthly | OECD | values |

| Industrial production index | Japan | monthly | OECD | services |
|---|---|---|---|---|
| Industrial production index | Mexico | monthly | OECD | volumes |
| Industrial production index | Russia | monthly | OECD | values |
| Manufacturers' new orders | United States | monthly | FRED | volumes, services |
| Manufacturing business activity confidence indicator | Poland | monthly | OECD | volumes |
| Manufacturing employment future tendency | Italy | monthly | OECD | services |
| Manufacturing export order books | Germany | monthly | OECD | values |
| Manufacturing export order books | Italy | monthly | OECD | services |
| Manufacturing export order books | United Kingdom | monthly | OECD | services |
| Manufacturing order books | Italy | monthly | OECD | services |
| Manufacturing order books | Netherlands | monthly | OECD | volumes |
| Merchandise exports | Brazil | monthly | OECD | values |
| Merchandise exports | Italy | monthly | OECD | values, volumes |
| Merchandise exports | Japan | monthly | OECD | values, volumes |
| Merchandise exports | Netherlands | monthly | OECD | values, services |
| Merchandise exports | South Korea | monthly | OECD | volumes |
| Retail trade index, values | France | monthly | OECD | services |
| Retail trade index, values | Spain | monthly | OECD | services |
| Retail trade index, values | United States | monthly | OECD | volumes |
| Retail trade index, volumes | France | monthly | OECD | values |
| Retail trade index, volumes | United States | monthly | OECD | volumes |
| Total air freight | Hong Kong airport | monthly | HKG | values |
| Total container throughput | Singapore | monthly | Singapore DOS | volumes |
| Total merchandise exports | global | quarterly | WTO | values, volumes, services |

## Appendix 2. Nowcast evolution over time, mean and ARMA ragged edges filling

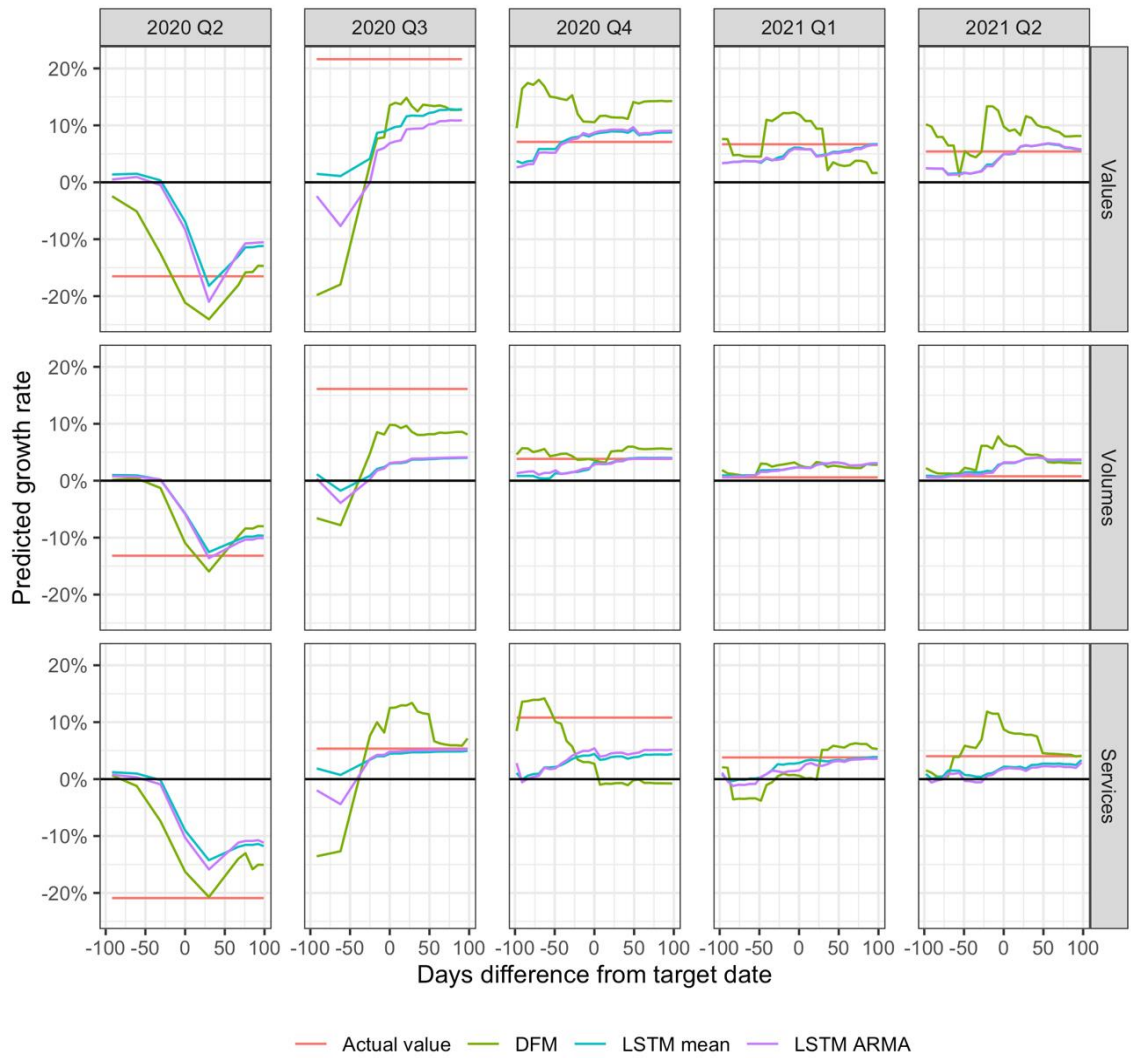